# The Dark Side of AI Transformers: Sentiment Polarization & the Loss of Business Neutrality by NLP Transformers

Prasanna Kumar
pk@businessoptima.com
Business Optima

## Abstract

The use of Transfer Learning & Transformers has steadily improved accuracy and has significantly contributed in solving complex computation problems. However, this transformer led accuracy improvement in Applied AI Analytics specifically in sentiment analytics comes with the dark side. It is observed during experiments that a lot of these *improvements in transformer led accuracy of one class of sentiment has been at the cost of polarization of another class of sentiment and the failing of neutrality*. This lack of neutrality poses an acute problem in the Applied NLP space, which relies heavily on the computational outputs of sentiment analytics for reliable industry ready tasks.

1. **Introduction**

Industry, Applied AI practitioners & Researchers have been using a plethora of NLP transformer technologies for processing sentiment. One of the key observation in Natural Language Processing using pre-trained models and transformers is that it produces excellent computational accuracies for i) language translation, ii) sentence prediction and iii) to a significant extent in natural language generation (NLG), however it performs poorly when it comes to sentiment computation. Models almost often require massive retraining and overhauling when it needs to be reliably applied for industry applications.

The paper, "Attention Is All You Need", (Vaswani et al., 2017), [1706.03762] proposes a landmark approach for a completely new type of model – The Transformer. Transfer Learning approach for neural network outlined in TransferTransfo [1901.08149] **(**Thomas Wolf et al, 2019) indicates the rapid adoption in conversational agents, which is a key applied AI requirement. Packed with pre-trained language models, Transformers as state of art natural language processing (Thomas Wolf, et al., 2020) [1910.03771] is pitted for a wide variety of tasks. With the coming together of pre-training, transfer learning and transformers, Industry did observe significant improvement in the field of machine translation and language generation for applied AI tasks. A paper on comprehensive survey of knowledge enhanced pre-trained language models (Xiaokai et al 2021) does validate this new paradigm. However*, the impact of pre-training, transfer learning & transformers on sentiment analytics is a mixed bag.*

Our primary goal is to unravel insights into why the bright side of transfer learning & transformer led Improvements in accuracy has a dark side of systemic polarization. This work goes beyond the usual suspects, such as datasets quality issues, biases and stereotypes, into the more subtle yet highly impactful role of transfer learning and transformer in sentiment polarization. This is our validation of whether systemic polarization exists across several generations of NLP Transformers. Is it observable independent of the dataset, groundings, biases & stereotypes? Is it more to do with the way transformers separate sentiments, given the analysis viewpoint that most sentiments, especially neutral sentiments, do not exist in the margins of strong sentiments (positive and negative). Could this be the reason why it takes a subsequent lot of training effort to depolarize the sentiment to get back the neutrality? The **loss of neutrality** and associated **polarity leakage** is a cause of concern for reliable rapid implementation of NLP transformers for industry ready application. I look forward to presenting our findings, analysis and answers for these questions.

## 2. Related Work

Several papers points out systemic lacune & limitations in the following area a) word-embeddings' sentiment mis-representations in tensor subspaces (Thomas Manzini et al, 2019) b) lack of grounding truth in annotations on technical content as hurdles (Kalpesh Krishna et al., 2021) c) the poor correlation of popular metrics BLEU, ROUGE, METEOR, NIST, LEPOR, BERTScore, Greedy Matching, VectorExtrema with human judgment (Ananya et al., 2020) Even state of the art model are correlating poorly on human judgment as outlined in the survey of evaluation metrics (Ananya et al., 2020) d) absence of credible standardized test-bed for auto evaluation (Thomas Scialom et al., 2021) e) issue related to summarization not capturing multi-polarity sentiment and many more. As I advent into the era of Natural language generation (NLG) and associated use of sentiment analytics in applied AI for empathetic tasks, the impact of these sub-system lacunae is more evident.

*The focus of this paper however will be specifically on the effect of generations of NLP transformers on the loss of neutrality and the systemic polarization of sentiments in the applied AI tasks, resulting in dark side effect such as unsympathetic responses, hallucinatory response, exclusionary behavior, etc.*

Source attribution: Datasets, Figures in related work section, and certain textual references throughout the paper belong to the author(s) of those respective papers cited in tables, reference section. If any citation, source attribution is missing, it may kindly be understood as an oversight.

### 2.1 Comparing and Combining Sentiment Analysis Methods

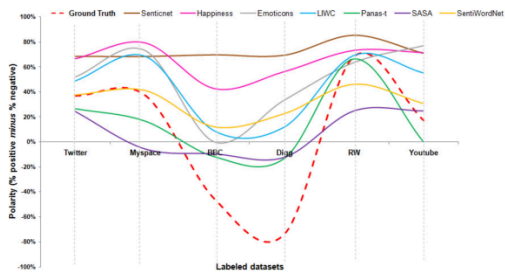

Figure: Polarity of 8 sentiments methods, labeled datasets

Pollyanna Gonçalves, et al 2014 outlined the presence of polarization across labeled datasets indicating the existing methods vary in agreement. The one possible explanation for the absymal performance of the NLP transformer I observed in our experiments came from the following paper by Xiaotao Gu et al.

### 2.2 Transformer Growth for Progressive BERT Training

Existing methods only conduct network growth in a single dimension, (Xiaotao Gu et al 2020). This paper outlined that it is beneficial to use compound growth operators and balance multiple dimensions (e.g., depth, width, and input length of the model).

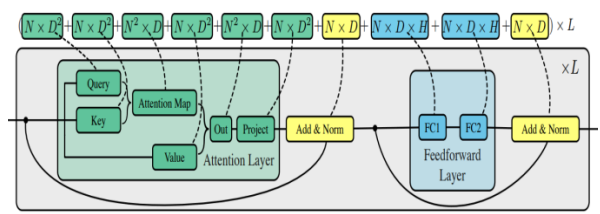

Figure: Cost of Different parts of Transformer Model

Due to the excessive cost of large-scale language model pre-training, considerable efforts have been made to train BERT progressively. BERT transformers by design start from an inferior, low-cost model and gradually grow the model to increase the computational complexity. (Xiaotao Gu et al.,2020)

So this model issue is systemic and by design. This is profound when it comes to emotions recognition. That brought us to analyze a paper which could compare performance of sentiment (emotion) recognition across multiple transformers.

### 2.3 Transformers in Emotion Recognition: a comparison of BERT, DistillBERT, RoBERTa, XLNet and ELECTRA

In sentiment analysis, F1 scores for emotion recognition, are pretty low across the spectrum. According to Diogo Cortiz, et al 2021, ELECTRA model, had the worst F1-score (.33), the other models had more similar results. RoBERTa achieved the best F1-score (.49), followed by DistillBERT (.48), XLNet (.48), and then BERT (.46). ELECTRA does not rely on masked language such as BERT or RoBERTa. Instead of masking the input and predicting it, ELECTRA replaces tokens with alternatives generated by a small generator network. So for the diversity in transformer mix, I have factored both BERT based transformers and ELECTRA based transformers for our experiments.

I also want to understand biases & stereotypes to mainly figure out if polarization has anything to do with the unreliable nature of transformers. Follow up with experiment to verify whether polarization does occur even on a completely unbiased human labeled ground truth. It would help showcase transformers' role in sentiment polarization. First I want to independently understand biases and de-biasing, any ineffectiveness & lacunae in de-biasing.

## 2.4 Usual Suspects : Biases, Stereotypes, Dataset, Embeddings & De-biasing

Online texts, across genres, registers, domains, styles, are riddled with human stereotypes" (Thomas Manzini et al, 2019), is outlined in the paper "Black is to Criminals, as Caucasian is to Police". Paper exclaims "in addition to possessing informative features useful for a variety of NLP tasks, word embeddings reflect and propagate social biases present in training corpora (Caliskan et al., 2017; Garg et al., 2018). Machine learning systems that use embeddings can further amplify biases (Barocas and Selbst, 2016; Zhao et al., 2017)."

**De-biasing Solution**
The multiclass de-biasing solutions offered (Thomas Manzini et al, 2019) are hard & soft de-biasing. Hard de-biasing, firstly required neutralize/equalize not-gendered words, secondly, gendered words embeddings are centered and their bias components are equalized eg: male bias and female bias must be equal magnitude. Soft De-biasing requires "learning a projection of the embedding matrix that preserves the inner product between bias and de-biased embeddings while minimizing the projection onto the bias subspace that should be neutral."

**Limitations**
The de-biasing methods based on bias component removal are insufficient to completely remove bias in the embeddings, (Gonen and Goldberg, 2019), since embeddings with similar biases are still clustered together after bias component removal. However, according to the author (Thomas Manzini et al), increasing the size of the bias subspace reduces the correlation of two variables and reduces the bias. Source: [1904.04047]

The following paper added additional information and relevant knowledge: Man is to Computer Programmer as Woman is to Homemaker? De-biasing Word Embeddings, Tolga Bolukbasi, et al, Source: [1607.06520]. Other authors whose papers provided insights in the related topic are Barocas & Selbst, 2016; Zhao et al., 2017

During our experiments I did not observe significant, consistent bias reduction across different datasets. As I analyzed this issue further, I landed on this paper by Po-Sen Huang, et al, 2020, which outlined that de-biasing of word embeddings to be quite ineffective. I needed a paper to validate and the following did address that requirement.

## 2.5 Evaluating sentiment bias in GPT-2
Po-Sen Huang, et al, 2020

In this paper, I found empirical sentiment distribution of text generated by GPT-2. The paper reveals systematic differences in sentiment depending on occupation. To quantify sentiment bias, the author proposes the use of individual and group fairness metrics from the fair machine learning literature (Dwork et al., 2012; Jiang et al., 2019; Hardt et al., 2016). Rather than debiasing word embeddings, Lu et al. (2018) proposed counterfactual data augmentation as a remedy to occupation-specific gender biases. The paper outlines the findings that **it can be much better to retain model performance** than debiasing word embeddings, especially in language modeling. Source: [1911.03064]

Few of the related work are 1: Language Models are Few-Shot Learners, Tom B. Brown, et al, 2020. GPT-3 in spite of its superior performance still has issues. GPT-3 faces methodological issues related to training on large web corpora. Source: [2005.14165]. Related work 2: Bertscore: Evaluating text generation with Bert. Tianyi Zhang, Varsha Kishore, Felix Wu,

Kilian Q Weinberger, and Yoav Artzi. It is necessary to both detect polarization and differentiate offensive words from polite complaining texts, so as to treat them accordingly.

## 2.6 Offensive Language and Hate Speech Detection

Understanding the relationship between polarity and offensive languages is an input for the work. "Differentiating if a text message belongs to hate speech and offensive language is a key challenge in automatic detection of toxic text content" (Bencheng Wei, et al., 2021). In short, "*Hate class usually has more negative sentiment (Polarity) compared to Neither and Offensive.*" Knowledge of Polarity helps differentiate hate class from offensive and neither class. So it is critical that a few more examples perhaps are worth more than billions of parameters (Yuval Kirstain, et al., 2021). Related work: Beyond Accuracy: Behavioral Testing of NLP models with Checklist (Marco et al 2020).

*The polarization & loss of neutrality I observed both in our applied AI projects and in research experiments. There were a significant number of papers on bias and stereotypes but not on sentiment polarization or the loss of neutrality. This was indeed a key motivator for us to take up the task of recording our observations & writing a paper.*

## 3. Limitations

Key limitations are 1) Experiment Results Merge Issues: The difficulty of merging of results coming from different experiments on disparate sub systems technology 2) Metric Issues: Non Translatability of F1 Macro to the applied AI space covering business rules. Lack of more Direct Correlation metrics for Darkside impact because of business rules of economical distribution of CSR Ticketing time & CSR Bot resources, 3) Lack of Pipeline Neutrality, Metric Transparency 4) Factual Inconsistencies 5) Model Hallucination. 6) Subjectivity of Decisions in Applied AI: the decision to act upon (Call to Action) based on sentiment classification analysis is multi-layered & subjective. Whether a customer should be responded to, attended to, should a ticket be raised is dependent on economics of time, budget availability & workloads. In addition, there is the emerging trend of Social Media Policing & Conformance requirements resulting in loss of shrinking the space for truthful sentiment expression, loss of diversity of opinions and emergence of monotonic predictably polarize-able sentiments. Existing F1 Macro has limitations in measuring these as it requires a multi-disciplinary metric approach.

4. **Data**

Given below is the tabulated description of the dataset(s) that the project will use for evaluation.

### 4.1 Datasets & Pre-Trained Embeddings

| # | Dataset | Source & Description |
|---|---------|----------------------|
| 1. | Twitter US Airline Sentiment Dataset 2015: (ternary sentiment) | - https://www.kaggle.com/crowdflower/twitter-airline-sentiment<br>- Original source: https://appen.com/open-source-datasets/<br>- 14000 record |
| 2. | Customer Support dataset with Call-to-Action | - https://www.kaggle.com/thoughtvector/customer-support-on-twitter<br>- Random Sampling of 3 million records |
| 3. | Baseline & Gold References | - Hand labeled by authors with Applied AI business insight, validation records - 18<br>- Baseline: Vader Sentiment |
| 4. | Pre-trained embeddings : | - BART Eli5 - yjernite/bart_eli5<br>- BERT_Base_uncased<br>- distilbert-base-uncased-finetuned-sst-2-english<br>- roberta-large<br>- Electra-small-discriminator |

Table 1: Dataset & Pre-Trained Embeddings

### 4.2 Ground Truth

| Complaining Text | Ground Truth |
|------------------|--------------|
| My flight got delayed, you guys suck, worst airline | negative |
| My flight got delayed, this is fxxxing great | negative |
| My flight got delayed, somebody is going to get hurt, real bad | negative |
| My flight got delayed, really bad service | negative |
| My flight got delayed, if I don't get this fixed, you are fired | negative |
| My flight got delayed, you guys better resolve this ASAP | negative |
| My flight got delayed, Need an explanation | neutral |
| My flight got delayed, Need an resolution | neutral |
| My flight got delayed, I need help | neutral |
| My flight got delayed, Need an response | neutral |
| My flight got delayed, Need rescheduling | neutral |
| My flight got delayed, Need an new ticket | neutral |
| My flight got delayed, This is fantastic, Hope help will arrive | positive |
| My flight got delayed, there should be a good reason | positive |
| My flight got delayed, It is still ok | positive |
| My flight got delayed, yet we arrived on time | positive |
| My flight got delayed, I'm glad I landed on time | positive |
| My flight got delayed, there is a good reason | positive |

Table 2: Collection of similar & differentiated complaining text across 3 sentiment classes

5. **Metrics**

The metrics that will form the basis for evaluation are F1 Macro for standard baseline evaluation across multiple transformers. I also used a more focused F1 Macro of Neutral Class for additional insights. Along with neutral specific F1 Macro, I also use F1 Macro for positive and negative sentiment class. This I found was most appropriate given the scope was focused on neutrality and sentiment polarization.

For future work, these are some of the metrics I propose would be useful to measure loss of neutrality at various stages inside the NLP pipeline.

- F1 Neutral_Precision
  (Measure Pipeline Neutrality / Neutral_Precision)
- F1 Neutral_Loss_Type_1
- (Neutral_to_Negative Loss inside the Pipeline)
- F1 Neutral _Loss_Type_2
- (Neutral_to_Positive Loss inside the Pipeline)

Measuring and side by side reporting of neutral mis-classification both with and without gold reference across the NLP pipeline. This reporting method will add tremendous transparency to the industry wide rapid adoption of NLP pipeline transformation. Transformers could be better benchmarked with this method for industry adoption.

6. **Models**

Transformer base models used for evaluation are BERT, DistilBERT, RoBERTa, ELECTRA. VADER Sentiment is used for Lexicon & Rule based Baseline Reference. These models & tools are from respective AI implementation providers. Human judgment is assessed using hand labeled ground truth and validation sampling from dataset sources mentioned in the data section. Credit: "Hugging Face" models (https://huggingface.co/transformers/)

7. **Experiments & Results**

| Dark-Side Experiments | Description |
|---|---|
| 101 | Establish clear presence of Sentiment polarization in models |
| 102 | Transformer Growth, Transfer Learning & Stereotype Distribution<br>- Transfer Learning from Pre-Trained Models, Dataset Inheriting & amplifying polarizations |
| 103 | Conduct experiments on lack of Sentiment Comprehension<br>- factual Inconsistencies<br>- classification failures |
| 104 | Conduct metrics drill down experiments<br>- Compute F1 Macro<br>- Compute F1_Neutral & Polarized Class. Observe if positives have a pass, negatives filtered down or responded, neutrals ignored |
| 105 | Conduct experiments on auto evaluation and factual consistencies on human judgment. harness sentiment auto evaluation capability |
| 106 Future Work | Other systemic hurdles to progress<br>- More detailed experiments on hallucinatory response<br>- Investigate the growing influence of social media policing & policy conformance<br>- shrinking space truthful expression |

Table 3: List of Dark-Side Experiments

These pointed experiments are designed to bring out the dark side aspect of polarization, hallucination and exclusion. Experiments leverage the existing F1 macro metric to pin point class polarization. Demonstrate hallucination from pre-trained off the shelf transformers. Finally, establish the dark side of transformers in customer service where in usage transformer polarization accentuates the impact of offensive language by scoring them high, while ignoring complaining text that has polite language. Here are the results of transformer classification across various generations along with ground truth. Vader is used as an additional benchmarking tool that can work off the shelf tool. It is critical for us to show transformers issues against both these benchmarks.

| Ground Truth | Vader (Lexical based) | BERT (Transformer based) | DistilBERT (Transformer based) | RoBERTa | ELECTRA (Transformer based) |
|---|---|---|---|---|---|
| negative | negative | negative | negative | negative | negative |
| negative | neutral | negative | positive | negative | negative |
| negative | neutral | negative | negative | negative | negative |
| negative | negative | negative | negative | negative | negative |
| negative | neutral | negative | negative | negative | negative |
| negative | neutral | negative | negative | negative | negative |
| neutral | neutral | neutral | negative | negative | negative |
| neutral | neutral | negative | negative | negative | negative |
| neutral | neutral | neutral | negative | negative | negative |
| neutral | neutral | neutral | negative | negative | negative |
| neutral | neutral | negative | negative | negative | negative |
| neutral | neutral | negative | negative | negative | negative |
| positive | positive | negative | positive | negative | negative |
| positive | neutral | negative | negative | negative | negative |
| positive | neutral | neutral | positive | negative | negative |
| positive | neutral | negative | positive | negative | negative |
| positive | neutral | negative | positive | negative | negative |
| positive | neutral | negative | positive | negative | negative |

Table 4: Sentiment Classification Results for Models

|  | Grounding | Vader | BERT | DistilBERT | RoBERTa | ELECTRA |
|---|---|---|---|---|---|---|
| negative(%) | 33.0 | 11.0 | 100.0 | 67.0 | 100.0 | 100.0 |
| neutral(%) | 33.0 | 83.0 | 0.0 | 0.0 | 0.0 | 0.0 |
| positive(%) | 33.0 | 6.0 | 0.0 | 33.0 | 0.0 | 0.0 |

Table5: Polarization (%), Validation Samples

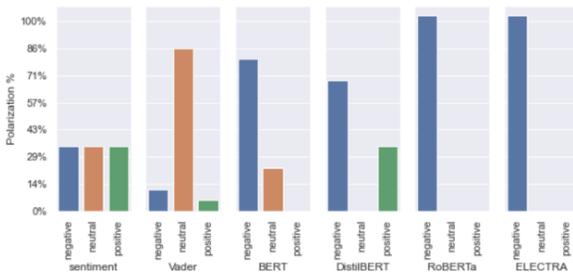

Figure1: Loss of neutrality, Validation samples

|  | Grounding | Vader | BERT | DistilBERT | RoBERTa | ELECTRA |
|---|---|---|---|---|---|---|
| F1_Macro | 1.0 | 0.452381 | 0.400000 | 0.462963 | 0.166667 | 0.166667 |
| F1_Macro_Neutral_Score | 1.0 | 1.000000 | 0.333333 | 0.000000 | 0.000000 | 0.000000 |
| F1_Macro_Positive_Score | 1.0 | 0.142857 | 0.000000 | 0.454545 | 0.000000 | 0.000000 |
| F1_Macro_Negative_Score | 1.0 | 0.250000 | 1.000000 | 0.454545 | 1.000000 | 1.000000 |

Table6: F1 Macro Score, & Class-wise F1 Macro Scores

|  | Grounding | Vader | BERT | DistilBERT | RoBERTa | ELECTRA |
|---|---|---|---|---|---|---|
| negative(%) | 82.0 | 1.0 | 85.0 | 80.0 | 100.0 | 83.0 |
| neutral(%) | 8.0 | 91.0 | 6.0 | 0.0 | 0.0 | 8.0 |
| positive(%) | 10.0 | 8.0 | 9.0 | 20.0 | 0.0 | 9.0 |

Table7: Polarization (%), Validation Samples (~300)

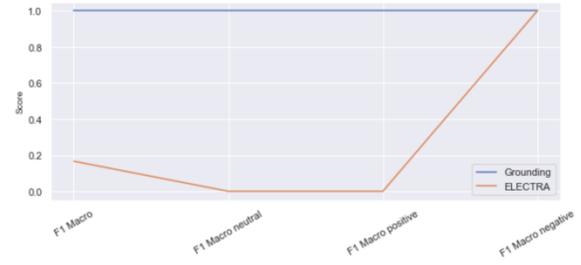

Figure2: Class-wise F1 Macro, Loss of Sentiment Neutrality (Ground Truth Vs Electra)

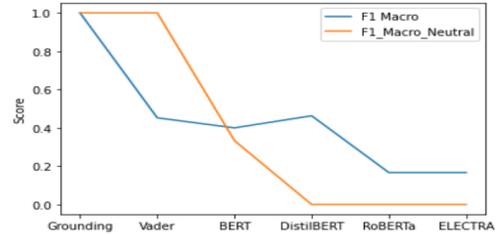

Figure3: Neutral Sentiment Loss, Polarization

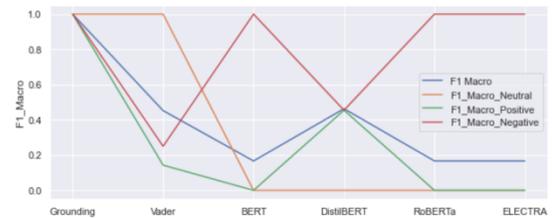

Figure4: F1 Macro of DistilBERT comparable with Vader

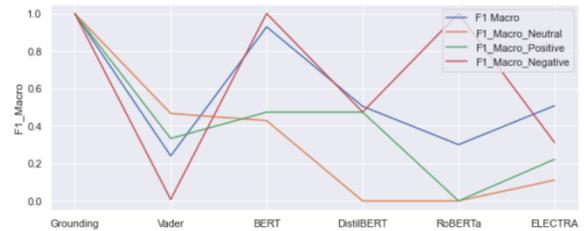

Figure5: Randomized Sub Sample of 1000 records

|  | Grounding | Vader | BERT | DistilBERT | RoBERTa | ELECTRA |
|---|---|---|---|---|---|---|
| F1_Macro | 1.0 | 0.240355 | 0.928849 | 0.504527 | 0.300366 | 0.508014 |
| F1_Macro_Neutral_Score | 1.0 | 0.466667 | 0.428571 | 0.000000 | 0.000000 | 0.111111 |
| F1_Macro_Positive_Score | 1.0 | 0.333333 | 0.473684 | 0.473684 | 0.000000 | 0.222222 |
| F1_Macro_Negative_Score | 1.0 | 0.008032 | 1.000000 | 0.474359 | 1.000000 | 0.311688 |

Table8: F1 Macro Class Scores for Vader & Transformers

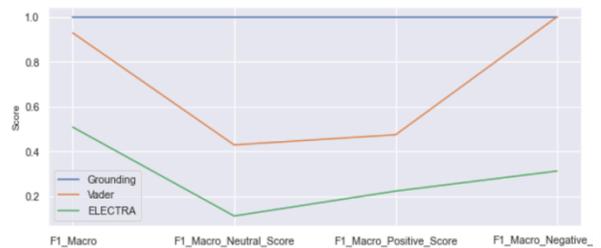

Figure6: F1 Macros Class Scores (Ground Truth, Vader, ELECTRA)

## 8. Analysis

In-spite of groundings & validation set having their classes balanced, with equal polarity for positive, negative and neutral, Transformer is still polarizing it into negative labels with high score. From Table [5] & Figure [1] there is a clear evidential proof for loss of neutrality and polarization of sentiments. As I increased the size of our training dataset, from Table [8] & Figure [5] I did observe improvement in Transformer F1 Macro Score. However the neutral classification still suffered no matter what. I have observed this trend with every increase in dataset size. There is a progressive increase in F1 Score of non-neutral class, and continued abysmal performance of neutral class.

In a mix of ablation study and counter factual analysis mode, when the same grounding dataset is used but without the use of a transformer, the results were incriminating. Even on the same sample size, a lexicon based Vader did much better, and is able to produce better F1 Score.

In processing sentiments on customer complaints dataset, which has a negative tonality, the neural and positive classes of Vader do suffer, but not as much as Transformer based models. These NLP transformers by default, start with the polarizing effect. Hence expensive training is required to depolarize them, with increasing training runs to hide the systemic drawbacks without addressing it.

It is observed that neutrals belong to different vector subspace(s) as compared to positives and negatives and not on their margins. What the transformer is doing is Gordian cutting of neutral and classifying them as positive and negative. Result is the accuracy improvement at the cost of wrongly classifying neutral sentiments. DistilBert neutrality loss is still very high compared to the ground truth and Vader benchmark. Following cases further reveal other issues accompanying NLP transformers

1. Pre-trained NLP Transformers prone to hallucination takes the model *from the bright-side of transfer learning into the dark-side of transfer hallucination*
2. Misclassification affecting workflow tasks
3. Systemic dwindling in neutral support
4. Offensive words are picked by automated CSR Bots for response as compared to polite words

**Case I: Sentiment, Summary & Hallucination**
Customer tweet = "really bad."
`Sentiment (polarity=-0.69,subjectivity=0.66)`

Unsympathetic Response Summary (Transformer):
`CSR Bot: We understand the issue you are facing is "really bad".` **I'm not a doctor, but I'm pretty sure it's not a good idea to drink a lot of alcohol.**
Text in Red color indicates hallucination of the transformer, subsequent summarization by customer service Bot.

**Case II: Adding Sentiment Analytics for Empathetic Response**
Customer tweet = "My flight is delayed, I need a good explanation"
`{'neg': 0.161, 'neu': 0.593, 'pos': 0.246, 'compound': 0.25}`

If Vader sentiment is plugged in the response
`CSR Bot:(Neutral Sentiment)  --no response--`

There is no system response as the neutral score is 0.593, followed by positive score, 0.246 that is higher than negative score 0.161.

If Transformer sentiment is plugged in the response
`[{'label': 'NEGATIVE', 'score': 0.99962258338}]`

Negative score is highly polarized. Resultant NLG summary response hallucinates as given below.

`CSR Bot:(Negative Sentiment)  I'm sorry you feel this way. We understand the issue you are facing is "My flight is delayed, I need a good explanation".` **It's not that my flight is delayed, it's that the flight is *delayed*. If you're going to be delayed, you have to be *very* delayed. If you don't, you're not going to make it**

Transformers should not be polarizing and hallucinating in the first place. These transformer models are now required to spend lot of development time to depolarize the model, and QA efforts to check for hallucination, and make it usable for production grade industry applications.

**Case III: Offensive Language for Attention Seeking, Response & Issue Resolution**

CSR data from multiple enterprises when used for training reveal, having one more offensive word, the more offensive the words, more Rating is secured, more Re-tweets it secures & as part of the Influencer management these critical inputs. As the transformer polarizing effect starts marking tweets and complaining text towards select classes, sentiment based analytics, let's say negative and positive, in order to keep up with response, customer service bot manager starts rationing and filters out mildly negative tweets and starts focusing on strong sentiments. That is, mildly upset tweets are being ignored and madly upset and antagonistic tweets are given more attention re-tweets, Direct Messaging requests and issue resolutions response.

```
Complaining Text (Polite):
[1] Please advise how to get compensation for
lost bag, i have spent far too long chasing
this
[2] how do you propose we do that
[3] Any update, Sophie? Thanks.
[4] Any chance you could help us out…

Did the CSR / CSR_Bot Respond: No

Complaining Text (Harsh):
[1] where are you f__ker
[2] please make this bullshit stop happening
[3] Go f__k yourselves, you greedy bas__rds
[4] fix this damn glitch

Did the CSR / CSR_Bot Respond: Yes
```

## 9. Future Work

I foresee our future work is likely to adopt knowledge graphs, innovative metrics & classification models to study and counter the ill effects of "Dark side of AI". In addition the roadmap could also include an integrated approach to sentiment analysis and semantic analysis for the applied AI tasks. It is a complex problem that requires further in-depth analysis, and is an excellent candidate for future work. In the spirit of openness & knowledge dissemination, I encourage AI communities to conduct additional research on this topic. I hope our work draws interest to this topic and dark-side of AI will get its due research attention.

## 10. Conclusion

Our experiments validate the presence of systemic polarization across different generations of transformers resulting in 1) High levels of sentiment polarization towards a certain class of sentiment. [Table 4,5,6] 2) Significant loss of neutrality [Figure 1,2]  3) Sentiment classification contradiction between models & benchmarks [Figure 1 & 5] 4) Transformer Model based CSR bot Hallucination [case-I, case-II] 5) Penalty for Politeness [case III]: If Enterprise Customer Support Bots have to be pick up limited number of complaint based on negative polarity scores, transformer polarization results in picking up complaining text with harsh languages rather than complaining text with polite language. Subsequent Exclusion of the polite sentiment with priority for lack of harsh negative language, implying "if you want customer support bot response & issue resolution then speak polarizing harsh language"

In summary, since certain classes, especially neutrals, don't exist in the margin of other classes, such as positives and negatives, transformer led separation results in polarization. This transformer led polarization led to excess activation of the sentiment analytics listener overloading the downstream sentiment consuming task (eg: ticketing systems, CSR bot, etc) with too many majority-class classifications. This results in many polite requests getting submerged & ignored. Transformers inadvertently accentuate the dark side effect by qualifying harsh language for response & ignoring polite complaining text. Successive generations of NLP Transformers, armed with transfer learning, have led us on a pathway of instant pre-trained polarization and loss of neutrality, requiring a lot of effort to depolarize. Indicating perhaps us researchers, industry practitioners collectively have been chasing a wrong set of transformer accuracy metric(s) relentlessly, while ignoring its dark side impact on Sentiment Analytics.

## 11. Authorship

- Prasanna Kumar Concept Development, Research, Conducting Experiments and Coding with Vader Sentiment, DistilBERT, RoBERTa, Grounding References, Authoring & Updates.